\pdfoutput=1
\documentclass[11pt]{article}

\usepackage[]{acl}

\usepackage{times}
\usepackage{latexsym}

\usepackage[T1]{fontenc}
\usepackage[utf8]{inputenc}

\usepackage{microtype}

\usepackage{inconsolata}
\usepackage{graphicx}
\usepackage{bm}
\usepackage{amsmath}
\usepackage{amsfonts}
\usepackage{xspace}
\usepackage{multicol}
\usepackage{multirow}
\usepackage{subfigure}
\usepackage{makecell}
\usepackage{listings}
\usepackage{adjustbox}
\usepackage{booktabs}
\usepackage{hyperref}
\usepackage{pifont}
\usepackage{enumitem} 

\usepackage[ruled, vlined, linesnumbered]{algorithm2e}

\definecolor{codegreen}{rgb}{0,0.6,0}
\definecolor{codegray}{rgb}{0.5,0.5,0.5}
\definecolor{codepurple}{rgb}{0.58,0,0.82}
\definecolor{backcolour}{rgb}{0.95,0.95,0.92}

\lstdefinestyle{mystyle}{
    backgroundcolor=\color{backcolour},   
    commentstyle=\color{codegreen},
    stringstyle=\color{codepurple},
    basicstyle=\ttfamily\scriptsize,
    breakatwhitespace=true,         
    breaklines=true,                 
    captionpos=b,                    
    keepspaces=true,                 
    numbers=none,                    
    numbersep=5pt,                  
    showspaces=false,                
    showstringspaces=false,
    showtabs=false,                  
    tabsize=2,
    columns=flexible,
    escapeinside={(*}{*)},
}

\newcommand{\ourdata}{\textsc{ImplexConv}\xspace}

\newcommand{\ourmodel}{\textsc{TaciTree}\xspace}

\title{Toward Multi-Session Personalized Conversation: A Large-Scale Dataset and Hierarchical Tree Framework for Implicit Reasoning}

\author{
Xintong Li, Jalend Bantupalli, Ria Dharmani, Yuwei Zhang, Jingbo Shang \\
University of California, San Diego \\
\texttt{\{xil240, jbantupalli, rdharmani, yuz163, jshang\}@ucsd.edu}\\
}

\begin{document}
\maketitle

\begin{abstract}
There has been a surge in the use of large language models (LLM) conversational agents to generate responses based on long-term history from multiple sessions.
However, existing long-term open-domain dialogue datasets lack complex, real-world personalization and fail to capture implicit reasoning—where relevant information is embedded in subtle, syntactic, or semantically distant connections rather than explicit statements. 
In such cases, traditional retrieval methods fail to capture relevant context, and long-context modeling also becomes inefficient due to numerous complicated persona-related details.
To address this gap, we introduce \ourdata, a large-scale long-term dataset with 2,500 examples, each containing approximately 100 conversation sessions, designed to study implicit reasoning in personalized dialogues.
Additionally, we propose \ourmodel, a novel hierarchical tree framework that structures conversation history into multiple levels of summarization. 
Instead of brute-force searching all data, \ourmodel enables an efficient, level-based retrieval process where models refine their search by progressively selecting relevant details.
Our experiments demonstrate that \ourmodel significantly improves the ability of LLMs to reason over long-term conversations with implicit contextual dependencies.
\end{abstract}

\section{Introduction}
\begin{figure}[t]
    \centering
    \includegraphics[width=1.0\linewidth]{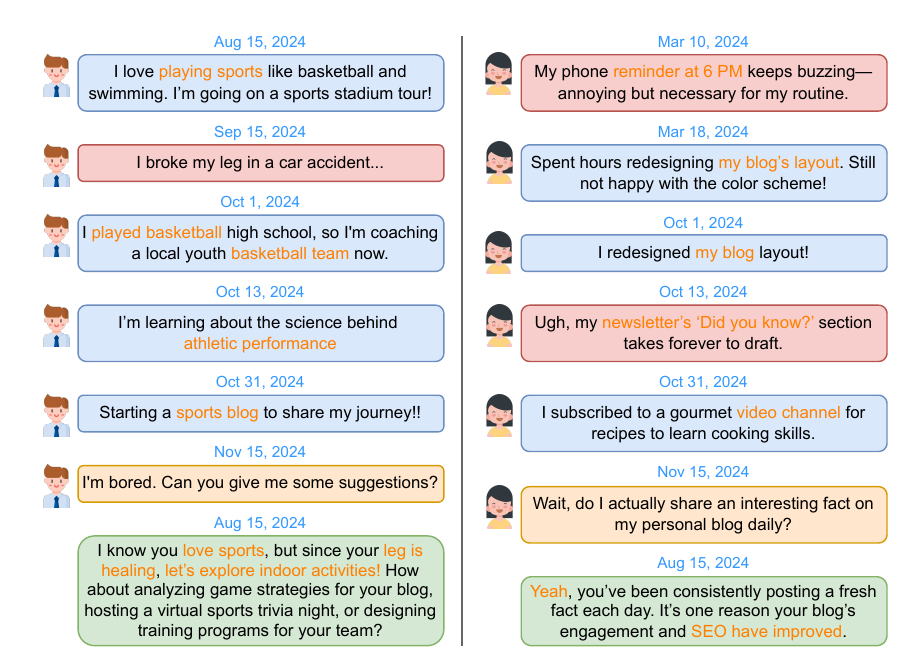}
    \vspace{-6mm}
  \caption{An example from \ourdata illustrating \emph{opposed} (left) and \emph{supportive} (right) implicit reasoning. 
  The orange block is the user query, the red blocks are implicit scenarios with low semantic similarity to the query, and the blue blocks are noisy but lexically related conversations that obscure the correct response.}
    \label{fig:implic_example}
    \vspace{-3mm}
\end{figure}

Large language models (LLMs) have revolutionized conversational AI by enabling personalized and context-aware dialogue generation~\cite{achiam2023gpt, mctear2022conversational}. 
Recent advances allow LLM-based agents to recall and integrate a long-term conversational history across multiple sessions, significantly enhancing coherence and personalization~\cite{zhong2024memorybank, li2024hello, wang2023enhancing}.
In this paper, we focus on \emph{implicit reasoning}, arguably the most challenging conversational setting, where relevant information is embedded in subtle syntactic patterns or semantically distant connections rather than explicitly stated, as demonstrated in Figure~\ref{fig:implic_example}.

As shown in Table~\ref{tab:dataset_comparison}, none of the existing datasets incorporates implicit reasoning scenarios --- Large-scale datasets~\cite{jang2023conversation} lack session depth, while deep but small datasets~\cite{wu2024longmemeval} lack structured personas critical for conversational consistency.
To bridge this gap, we construct \ourdata, a large-scale dataset with 2,500 multi-session examples ($\sim$100 sessions each) and 600 thousand persona traits designed to maintain session coherence.
\ourdata uniquely introduces implicit reasoning scenarios, where persona traits subtly reinforce or oppose other personalization details while maintaining low semantic similarity, making them difficult to trace. 

Implicit reasoning is particularly challenging to existing retrieval~\cite{shuster2021retrieval, fan2024survey} and long-context modeling~\cite{xiong2023effective, xu2023retrieval} techniques because it requires models to move beyond surface-level pattern recognition toward deeper reasoning over long-term interactions.
As dialogue history accumulates, numerous persona-related details can obscure critical implicit knowledge, making it increasingly difficult for long-context modeling techniques to extract and utilize relevant information effectively. 
The presence of excessive persona details often leads to retrieval inefficiencies, where dominant but less relevant traits overshadow essential implicit patterns, resulting in inconsistencies in generated responses.

\begin{table}[t!]{
\centering
\small
\resizebox{\linewidth}{!}{
\setlength{\tabcolsep}{1.5pt}{
\begin{tabular}{lcccc }
\toprule
\multirow{2}{*}{\textbf{Dataset}} & \textbf{\# of Conv.} & \textbf{Avg. Turns}  & \textbf{Avg.} & \textbf{Implicit}\\ 
  & / \textbf{\# Sessions} & \textbf{per Conv.}  & \textbf{Tokens} &\textbf{Reason} \\ 
\midrule
Daily Dialog~\cite{li2017dailydialog} & 13K / 13K & 7.9 & 114.7 & \ding{55} \\
PersonaChat~\cite{zhang2018personalizing} & 10K / 10K & 14.8 & 245.2 & \ding{55}\\
MSC~\cite{xu2021beyond} & 4K / 12K & 53.3 &  1,225.9 &  \ding{55}\\
% Conversation Chronicles~\cite{jang2023conversation} & 200K / 1M &  58.5 & 1,054.7  & \ding{55}\\
CC~\cite{jang2023conversation} & 200K / 1M &  58.5 & 1,054.7  & \ding{55}\\
LoCoMo~\cite{maharana2024evaluating} & 10 / 1K & 304.9 &  9,209.2  & \ding{55} \\
PerLTQA~\cite{du2024perltqa}  & 1  / 4K & 15K & 1M  & \ding{55} \\
\textsc{LongMemEval}~\cite{wu2024longmemeval} & 500 / 50K  & 5K & 115K & \ding{55} \\
\textbf{\ourdata (Ours)} & 2500 / 255K & 2K & 60K & \ding{51}\\
\bottomrule
\end{tabular}
}
}
\caption{Comparison of \ourdata with existing datasets, highlighting its large-scale multi-session structure and unique focus on implicit reasoning.}
\vspace{-6mm}
\label{tab:dataset_comparison}
}
\end{table}

We propose \ourmodel, a novel framework designed to address the inefficiency of retrieving implicit knowledge in long-term conversations, as shown in Figure~\ref{fig:framework}.
While LLMs can inherently assess whether an implicit scenario relates to a query, brute-force retrieval~\cite{lin2009brute} that inspects every individual fact can achieve a high recall, however, it would suffer from prohibitive computational costs.
\ourmodel overcomes this by structuring conversational history into a hierarchical tree, where lower-level nodes capture fine-grained details and higher-level nodes aggregate these into abstract summaries.
% By grouping relevant information into subtrees, our framework enables models to first evaluate high-level summaries, skipping entire subtrees deemed irrelevant.
By grouping relevant information into subtrees, our framework enables subtree skipping by evaluating high-level summaries --- only when a summary is relevant does the model drill down into finer-grained details. 
This hierarchical approach reduces the search space by orders of magnitude compared to brute-force retrieval while retaining high accuracy, as LLMs leverage their inherent reasoning ability to navigate the tree. 

We evaluate \ourdata and \ourmodel via question-answering tasks. \ourdata exhibits 20\% lower semantic similarity between queries and ground-truth answers compared to existing datasets, reflecting its unique challenge of high implicitness. 
\ourmodel achieves 30\% higher retrieval accuracy than baselines (e.g., RAG, MemoryBank), which struggle with implicit reasoning unless retrieving excessive amounts of information. 
Notably, \ourmodel achieves this with 40–60\% fewer tokens, demonstrating efficient extraction of implicit knowledge without sacrificing precision.

Our contributions are summarized as: 
\begin{itemize}[nosep,leftmargin=*]
    \item We introduce \ourdata, a large-scale multi-session dialogue dataset specifically designed to evaluate implicit reasoning in long-term personalized conversations.
    \item We propose \ourmodel, a hierarchical tree-based framework that efficiently stores and retrieves long-term conversational history, enabling models to extract implicit knowledge with level-based retrieval.
    \item Experimental results demonstrate the high implicitness of our dataset and the significantly improved retrieval accuracy of our framework, achieved with a smaller retrieval token size.
\end{itemize}
Our dataset and source code can be obtained here~\footnote{\url{https://github.com/Kaylee0501/ImplexConv}}.
\section{Related Work} 
Long-term conversational AI research spans both dataset construction and memory-enhanced methodologies. 
Existing multi-session dialogue datasets primarily focus on continuity, personalization, or memory retention, but they lack the necessary complexity for implicit reasoning. 
While datasets such as MSC~\cite{xu2021beyond} and LoCoMo~\cite{maharana2024evaluating} incorporate structured long-term interactions, they do not explicitly model implicit reasoning. 
Similarly, methodologies for long-term memory, including structured memory mechanisms~\cite{zhong2024memorybank} and RAG frameworks~\cite{lewis2020retrieval}, aim to improve historical context utilization but struggle with implicit dependencies
Additional discussions on related datasets and methodologies are provided in Appendix~\ref{sec:relatedwork}.

\section{\ourdata Collection}\label{sec:dataset}
We introduce \ourdata, a large-scale dataset designed to evaluate implicit reasoning in long-term multi-session conversations.
It comprises 2,500 examples, each containing approximately 100 dialogue sessions. 
Unlike existing datasets, \ourdata incorporates both opposed and supportive reasoning cases, where relevant information is embedded in subtle, syntactic, or semantically distant connections rather than explicit statements.
These properties make \ourdata a challenging benchmark for evaluating retrieval-based and long-context models.
Our dataset construction consists of persona extraction, implicit reasoning generation, and multi-turn conversation formulation to ensure realism and diversity.

\subsection{Persona Extraction}
We extract a diverse set of personas $\mathcal{P}$ from Persona Hub~\cite{ge2024scaling}, where each persona consists of a single-sentence description of an individual’s demographics, career, personal goals, or daily activities. 
A sample persona is provided in Appendix~\ref{sec:persona}.
Since maintaining persona consistency is crucial for long-term conversational coherence, we focus on personas related to everyday life and hobbies. 
Rather than using their original free-text format, we prompt an LLM $M_1$ to standardize each persona trait into a structured format explicitly starting with ``This person....'' to ensure clarity and consistency across generated conversations.

\subsection{Implicit Reasoning}
Implicit reasoning is fundamental to \ourdata, as it introduces a layer of complexity beyond explicit factual retrieval. 
Unlike direct reasoning, implicit reasoning remains semantically distant from the original persona trait, making it more challenging to detect.
For each persona trait $p \in \mathcal{P}$, we generate 20 opposed reasoning scenarios $R_o$ and 20 supportive reasoning scenarios $R_s$ using $M_1$.
As illustrated in Figure~\ref{fig:implic_example}, opposed reasoning introduces situations that prevent the persona from engaging in $p$ (e.g., ``I broke my leg'' opposes ``I enjoy playing sports''). 
Supportive reasoning subtly reinforces $p$ (e.g., ``My newsletter's section takes forever to draft'' supports ``I share facts on a personal blog'').
To select high-quality reasoning scenarios from $R_o$ and $R_s$, we leverage an instruction-tuned text embedding model $E$ to obtain vector representations $E_o$ and $E_s$ for the generated reasoning, as well as $E_p$ for the original persona trait $p$. 
We then compute the semantic similarity between each reasoning and $p$. 
To ensure that the selected scenarios do not exhibit high semantic overlap with the original trait, we filter out reasoning instances with a similarity score above a predefined threshold $\beta$. 
For cases where the similarity score is close to $\beta$, we incorporate human verification to refine the selection. This process results in the filtered sets $R_o'$ and $R_s'$.
Following this, we apply additional selection criteria separately for opposed and supportive reasoning to ensure a diverse and representative dataset. 
To facilitate evaluation across different frameworks, we further construct corresponding question-answer tasks for each type of implicit reasoning.

\begin{figure*}[t]
    \centering
    \includegraphics[width=1.0\linewidth]{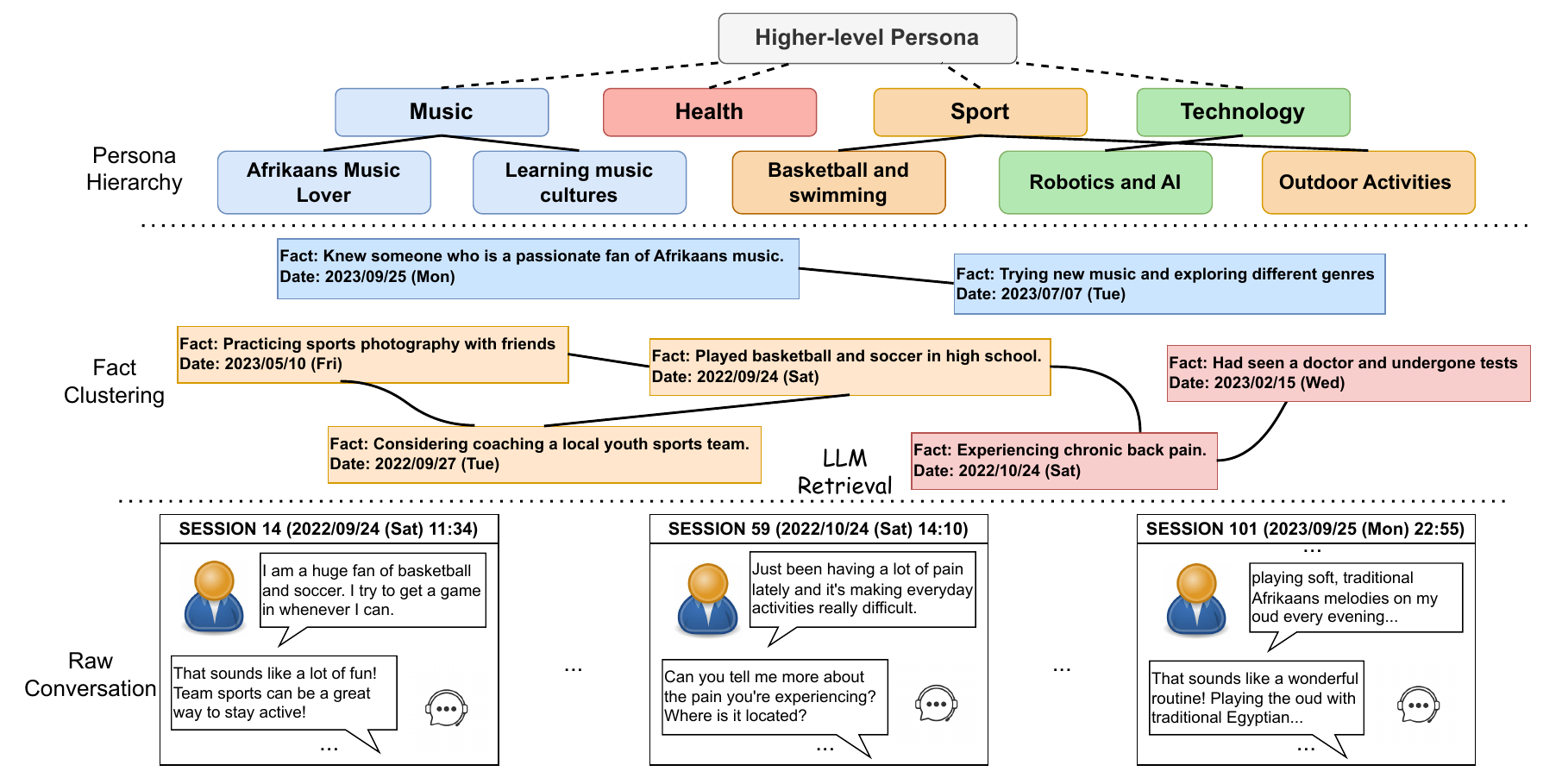}
    \vspace{-3mm}
  \caption{Overview of \ourmodel framework. \ourmodel organizes long-term conversational history into a hierarchical structure, clustering related facts to enable efficient retrieval of implicit reasoning. By leveraging LLMs to refine relevant information while discarding unrelated details, the framework reduces search space and improves retrieval efficiency.}
    \label{fig:framework}
    \vspace{-3mm}
\end{figure*}

\subsubsection{Opposed Reasoning Scenarios}
We aim to keep a single high-quality instance. 
Among $R_o'$, we prompt $M_1$ to identify the best reasoning instance $R_o^*$, ensuring that it is both implicit and strongly opposed to $p$. In cases where the LLM outputs multiple responses or fails to provide a suitable answer, human annotators manually select the most appropriate reasoning.
To construct the corresponding question-answer task, we generate a general question $q_o$ related to the persona trait, focusing on aspects of daily life. 
The question-answer pair must satisfy the following condition: Without $R_o^*$, the expected answer $a_o$ would align with $p$; however, due to the presence of the scenario, the person can no longer engage in $p$. $a_o$ must reflect this shift.

\subsubsection{Supportive Reasoning Scenarios}
Unlike opposed implicit reasoning, which presents a more challenging inference task, supportive implicit reasoning is designed to be comparatively easier to test.
Since the high-quality supportive scenario is hard to generate, for each reasoning instance in $R_s'$, we prompt $M_1$ again to verify its alignment with $p$.
To ensure reliability, we instruct the model to provide an answer only if it is certain; uncertain cases are manually reviewed by human annotators. 
The remaining verified reasoning instances, denoted as $R_s^*$, serve as the final supportive reasoning candidates.
The corresponding question-answer pairs are straightforwardly derived by transforming the trait into a yes/no question. The expected answer depends on whether the implicit reasoning supports the original persona trait.

\subsection{Conversation Formulation}
The final step in dataset construction involves generating conversations based on the previously curated scenarios.
Instead of simulating human-human dialogues, we design human-chat assistant interactions, which require tracking conversation history to generate appropriate responses. 
For opposed reasoning cases, we further increase the difficulty of identifying the implicit reasoning $R_o^*$. 
Specifically, we prompt $M_1$ to generate five additional scenarios related to $p$, each with higher semantic similarity to the target question $q_o$ than $R_o^*$. 
These scenarios serve as distracting context, making $R_o^*$ harder to detect.
The original persona trait $p$, the opposed implicit reasoning $R_o^*$, and these additional noisy scenarios are each expanded into multi-turn conversations between a human and an assistant, forming individual dialogue sessions. 
To ensure temporal coherence, we assign timestamps to each session, enforcing the constraint that $R_o^*$ does not occur immediately after $p$.
For supportive reasoning cases, conversations are generated based on all scenarios in $R_s^*$, with timestamps assigned accordingly.

To further enhance the realism of our dataset, we sample noisy sessions from real-world dialogue datasets, including CC~\cite{jang2023conversation}, LLM-Redial~\cite{liang2024llm}, and UltraChat~\cite{ding2023enhancing} for both cases. 
Since CC contains human-human interactions, we reformat them into a human-assistant dialogue structure. 
We randomly insert five sessions from each dataset with higher semantic similarity to $q_o$ and $q_s$ than $R_o^*$ and $R_s^*$, respectively. 
These evidence sessions are interleaved at random positions within the conversation history and assigned plausible timestamps to maintain consistency.
For supportive reasoning cases, we observe that some noisy sessions may directly reinforce $p$. 
To ensure high-quality selection, we employ both LLM-based verification and human review to filter out such instances.
As a result, each persona trait $p$ is associated with more than 15 corresponding sessions for both opposed and supportive reasoning cases. 
To scale the dataset, we randomly sample personas from $\mathcal{P}$ and merge their corresponding sessions, constructing \ourdata with approximately 100 sessions per example, resulting in a dataset of 2,500 instances.

\subsection{Implicitness of Datasets}
\begin{figure}[t]
    \centering
    \includegraphics[width=0.9\linewidth]{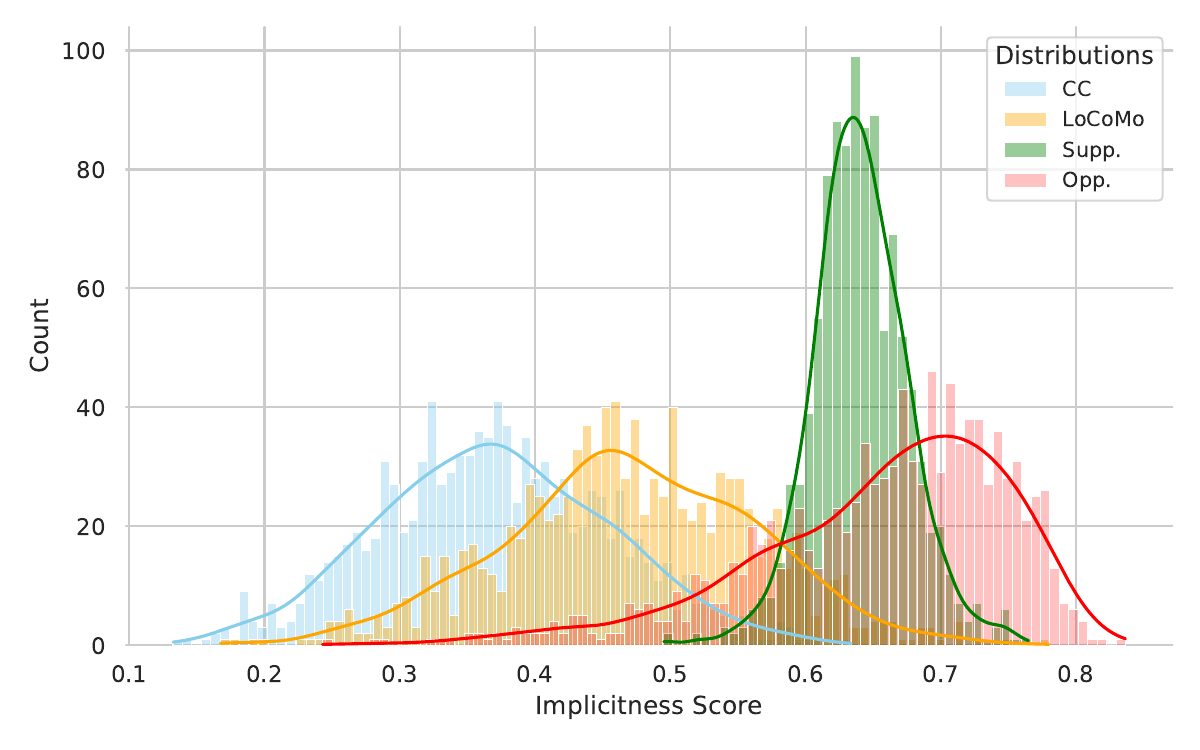}
    \vspace{-3mm}
  \caption{Distribution of implicitness scores across datasets, where Supp. and Opp. represent the supportive and opposed cases of \ourdata, respectively. }
    \label{fig:implicit}
    \vspace{-3mm}
\end{figure}

To evaluate the implicit nature of \ourdata, we measure the semantic similarity between the query and its corresponding answer across different datasets. A higher semantic distance indicates that the answer is less explicitly stated in the conversation history, making retrieval and reasoning more challenging. We define \textbf{Implicitness Score (IS)} as,
\begin{equation}
    IS = 1 - \text{Sim}(Q, A)
\end{equation}
where $Q$ is the query, $A$  is the ground truth answer, and Sim($\cdot$) represents cosine similarity over sentence embeddings.
Figure~\ref{fig:implicit} presents the distribution of IS across multiple datasets.
Traditional multi-session datasets, such as MSC and CC, exhibit relatively low implicitness scores, averaging around 37\%–0.38\%, indicating that their target information can often be retrieved through direct semantic similarity.
DailyDialog and LoCoMo display slightly higher scores, suggesting a moderate increase in reasoning complexity but still relying on explicit contextual cues.
In contrast, \ourdata demonstrates significantly higher IS scores, with supportive and opposed reasoning scenarios averaging 64\% and 65\%, respectively. 
These findings highlight \ourdata as a more demanding benchmark for evaluating retrieval-based models, requiring deeper inference beyond surface-level techniques.

\section{Framework}
We introduce \ourmodel, a framework designed to efficiently retrieve implicit reasoning from long-term conversational history. 
Unlike conventional retrieval methods that rely on direct semantic similarity, \ourmodel organizes historical information into a hierarchical structure, allowing for more effective navigation of implicit knowledge. 
Our key insight is that LLMs can recognize implicit reasoning relevance, but a brute-force approach—querying each fact individually—is computationally expensive. 
Instead, \ourmodel hierarchically clusters and summarizes history information, enabling efficient retrieval by progressively refining relevant information while discarding unrelated details. 
This structure allows retrieval to bypass irrelevant subtrees entirely, significantly reducing search space while preserving recall.
The framework overview is depicted in Figure~\ref{fig:framework}, and the below
sections provide a detailed discussion.

\subsection{Fact Extraction and Initial Clustering}\label{sec:cluster_trait}
Given a conversation session $c$ from the long-term history, we prompt an LLM $M_2$ to extract all facts that capture long-term conversational context, following the strategy used in \textsc{LongMemEval}~\cite{wu2024longmemeval}.
These extracted facts form a fact set $\mathcal{F}$, which often contains redundancy or excessive details. 
To improve efficiency, we apply a clustering technique to group semantically related facts while maintaining diversity.
We use the embedding model from Section~\ref{sec:dataset} to encode the extracted facts into vector representations. 
Then, we apply UMAP~\cite{mcinnes2018umap} for dimensionality reduction, followed by a Gaussian Mixture Model (GMM)~\cite{reynolds2009gaussian} for clustering, which offers flexibility in capturing complex data distributions. 
To ensure efficiency, each cluster is constrained to a maximum size $k$, leading to the total number of clusters at the initial level:
\begin{equation}
    H^0 = \left\lfloor \frac{|\mathcal{F}|}{k} \right\rfloor,
\end{equation}
where $H^0$ represents the number of clusters at the lowest level.
Each cluster contains related facts, which are then summarized by prompting $M_2$ to generate a condensed representation $s_i^0$ that retains all essential details. 
This set of summaries $\{s_i^0\}_{i = 1}^{H^0}$ serve as the leaf nodes of the hierarchical tree, forming the foundation for structured retrieval.

\subsection{Hierarchical Tree Construction}
With the fact summaries ${s_i^0}$ as leaf nodes, we construct a hierarchical tree through iterative clustering and summarization. 
At each hierarchical level $j$, we recursively group the summaries from the previous level:
\begin{equation}
    H^j = \left\lfloor \frac{H^{j-1}}{k} \right\rfloor.
\end{equation}
Each cluster at level $j$ is summarized into a higher-level abstraction $s_i^j$, providing progressively more general representations of the conversation history. The process continues until reaching the root-level size $L$, which represents the highest level of summarization.
%This hierarchical structuring is particularly effective for implicit reasoning retrieval. 
Instead of scanning all stored facts, \ourmodel enables LLMs to navigate from high-level summaries down to fine-grained details only when necessary. 
This approach reduces retrieval complexity and prevents noise from overwhelming the model, making it significantly more effective at capturing implicit reasoning.

\subsection{Information Retrieval}
\ourmodel enables efficient retrieval by leveraging its hierarchical structure. Given a query
$q$, we initiate retrieval at the highest level and progressively refine the search by navigating downward.
Instead of computing direct semantic similarity with all facts, we prompt $M_2$ at each level to identify relevant clusters: 
\begin{equation} 
S_{q}^j = { s_{i'}^j \mid M_2(q, s_i^j) \text{ is relevant} }. 
\end{equation}
This step drastically reduces search complexity, as irrelevant subtrees can be pruned early in the retrieval process.
For each selected summary $s_{i'}^j$, we continue the process recursively at level $j-1$, retrieving finer-grained summaries $S_{q}^{j-1}$ until reaching the leaf nodes $S_q^0$. At this stage, each retrieved summary $s_{i'}^j$ is linked to its corresponding original fact set $F_{i'}$, providing the final set of retrieved facts.
This hierarchical retrieval strategy ensures that only the most relevant implicit knowledge is selected while efficiently skipping unrelated information. 
Unlike brute-force approaches that inspect every individual fact, \ourmodel groups and filters information in a structured manner, significantly improving retrieval accuracy while maintaining efficiency.

\begin{table*}[t]
    \small
    \centering
    \renewcommand{\arraystretch}{1.2}
    \resizebox{\linewidth}{!}{
    \begin{tabular}{llcccccc}
        \toprule
        \multirow{3}{*}{\textbf{Category}} & \multirow{3}{*}{\textbf{Method}} & \multicolumn{6}{c}{\textbf{Datasets}} \\
        \cmidrule(lr){3-8}
        & & \multicolumn{4}{c}{\textbf{w/o} implicit reason} & \multicolumn{2}{c}{\textbf{w/} implicit reason}\\
        \cmidrule(lr){3-6} \cmidrule(lr){7-8}
        & & \textbf{MSC} & \textbf{CC} & \textbf{Daily Dialog} & \textbf{LoCoMo} & \textbf{IMPersona (Supp.)} & \textbf{IMPersona (Opp.)} \\
        \midrule
        \textbf{Memory-based}
        & MemoryBank & 17.78  & \textbf{39.24} & 24.51 & 10.97 & 15.90 & 7.95 \\
        \midrule
        \multirow{3}{*}{\textbf{RAG}} 
        & Raw  & 2.26 & 8.68 & 7.50  & 4.64 & 2.31 & 0.65\\
        & Summary & 15.73 & 38.39 & 34.03 & \textbf{26.75} & 12.66 & 5.58\\
        & GraphRAG & 5.00 & 5.27 & 11.70 & 10.28 & 3.00 & 0.85 \\
        \midrule
        \multirow{2}{*}{\textbf{\ourmodel}} 
        & Facts & 21.88 & 26.20 & 43.72 & 14.36 & 28.96 & 7.62 \\
        & Summary & \textbf{36.65}  & 37.70 & \textbf{55.86} & 16.63 & \textbf{55.18} & \textbf{14.84} \\
        \bottomrule
    \end{tabular}
    }
    \caption{Retrieval Accuracy (F1 score) across Different Frameworks and Datasets}
    \label{tab:retrieval_acc}
\end{table*}

\section{Experiments}
This section presents our experimental setup, including the datasets, baseline methods, evaluation metrics, and implementation details. We evaluate models on multi-session conversations using retrieval accuracy, answer correctness, and token efficiency, ensuring a rigorous comparison across different approaches.

\subsection{Experimental Settings}

\paragraph{Datasets.}
We conduct experiments on five benchmark datasets to evaluate how well models handle long-term history in multi-session conversations. 
These datasets include \textbf{DailyDialog}, \textbf{MSC}, \textbf{CC}, \textbf{LoCoMo}, and our proposed dataset, \textbf{\ourdata}.  Table~\ref{tab:dataset_comparison} provides an overview of each dataset’s information. 
The number of sessions per conversation ranges from 1 to 100, enabling a comprehensive evaluation of the models' ability to store and retrieve relevant information across different conversation lengths.

\paragraph{Compared Methods.}
We compare our \ourmodel framework with three types of baselines: memory-based methods, RAG approaches, and long-context models. For memory-based methods, we use \textbf{MemoryBank}~\cite{zhong2024memorybank}, which is designed to store and retrieve long-term persona information. 
MemoryBank simulates human memory retention by dynamically updating stored information over time.
RAG-based approaches retrieve relevant information using different selection strategies: a simple semantic \textbf{similarity-based} retrieval, a \textbf{summarization-based} approach that condenses raw conversations before retrieval, and \textbf{GraphRAG}~\cite{edge2024local}, which organizes knowledge into a structured graph and applies community detection for modular, query-focused summarization. 
Long-context models, in contrast, process full conversation histories without retrieval, either using the raw dialogue in its entirety or extracting key facts before feeding them into the model. 

\subsection{Evaluation Metrics}
There are various ways to evaluate baseline performance, including question answering, event summarization, and dialogue generation. 
In this paper, we focus on \textbf{question answering (QA) task}, as it provides a clear and intuitive way to assess both the implicit nature of \ourdata compared to other benchmark datasets and the performance of baseline frameworks.
We use three key metrics to evaluate the performance of different frameworks. 
\paragraph{Retrieval Accuracy.} 
To evaluate how well a framework retrieves the necessary background information while ensuring efficiency, we use the F1 score, which balances both relevance (recall) and conciseness (precision). 
Given a retrieved context $C_r$ and the ground truth context $C_g$, we define retrieval accuracy as,
\begin{equation}
\text{Retrieval Accuracy} = \frac{2 \times |C_r \cap C_g|}{|C_r| + |C_g|},
\end{equation}
where $|C_r \cap C_g|$ represents the overlap between the retrieved and ground-truth contexts. This formulation ensures that retrieved content is both comprehensive and efficient, avoiding excessive retrieval that may introduce noise.

\paragraph{Answer Correctness.} 
Since it is difficult for the predicted response to exactly match the ground truth, we prompt an LLM to judge whether the predicted answer is semantically equivalent to the ground truth. We also conduct human evaluation to verify correctness.
\paragraph{Token Efficiency.} 
While more information generally improves performance, excessive token usage increases computational cost. We analyze the trade-off between performance and token usage by measuring the token-to-accuracy ratio, which helps assess efficiency in long-term conversation retrieval.

\subsection{Implementation Details}
To mitigate hallucination, we employ different LLMs for dataset generation and framework implementation. 
Specifically, we use \texttt{Llama-3.1-405B-Instruct}~\cite{touvron2023llama} to construct the dataset and \texttt{GPT-4o-mini}~\cite{achiam2023gpt} to implement the framework and evaluate its performance across various baselines. 
For embedding representations, we use \texttt{stella\_en\_1.5B\_v5}~\cite{zhang2024jasper} as the embedding model $E$. All prompts used in this paper are provided in the Appendix.
To ensure diverse implicit reasoning scenarios, we set the similarity threshold $\beta = 0.4$. 
For clustering facts, we set the maximum cluster size to $k = 6$.  
The root-level cluster size $L$ is set to 15. 
If the number of nodes at a level drops below this threshold, we terminate clustering and designate it as the root level to preserve high-level summarization.

For the QA task, we evaluate model performance across both multi-session and single-session datasets to ensure consistency. We include DailyDialog, a single-session dataset, to verify that our framework does not degrade in simpler dialogue settings. 
Since MSC and CC do not provide explicit QA pairs, we treat the first four sessions as conversation history and evaluate QA performance based on responses in the fifth session.
Furthermore, because \ourdata comprises two distinct reasoning types—supportive and opposed implicit reasoning—each with fundamentally different QA dynamics, we evaluate them separately. 
To assess whether retrieved summaries or detailed facts contribute more effectively to accurate responses, we conduct evaluations using both the retrieved summaries $s_{i'}^j$ and their corresponding fact sets $F_{i'}$ independently.
\section{Results}
This section presents our results on retrieval accuracy, response accuracy, and token efficiency. We analyze how well models retrieve implicit information, the trade-off between accuracy and token usage, and the challenges of opposed implicit reasoning.

\begin{figure*}[t]
    \centering
    \includegraphics[width=1.0\linewidth]{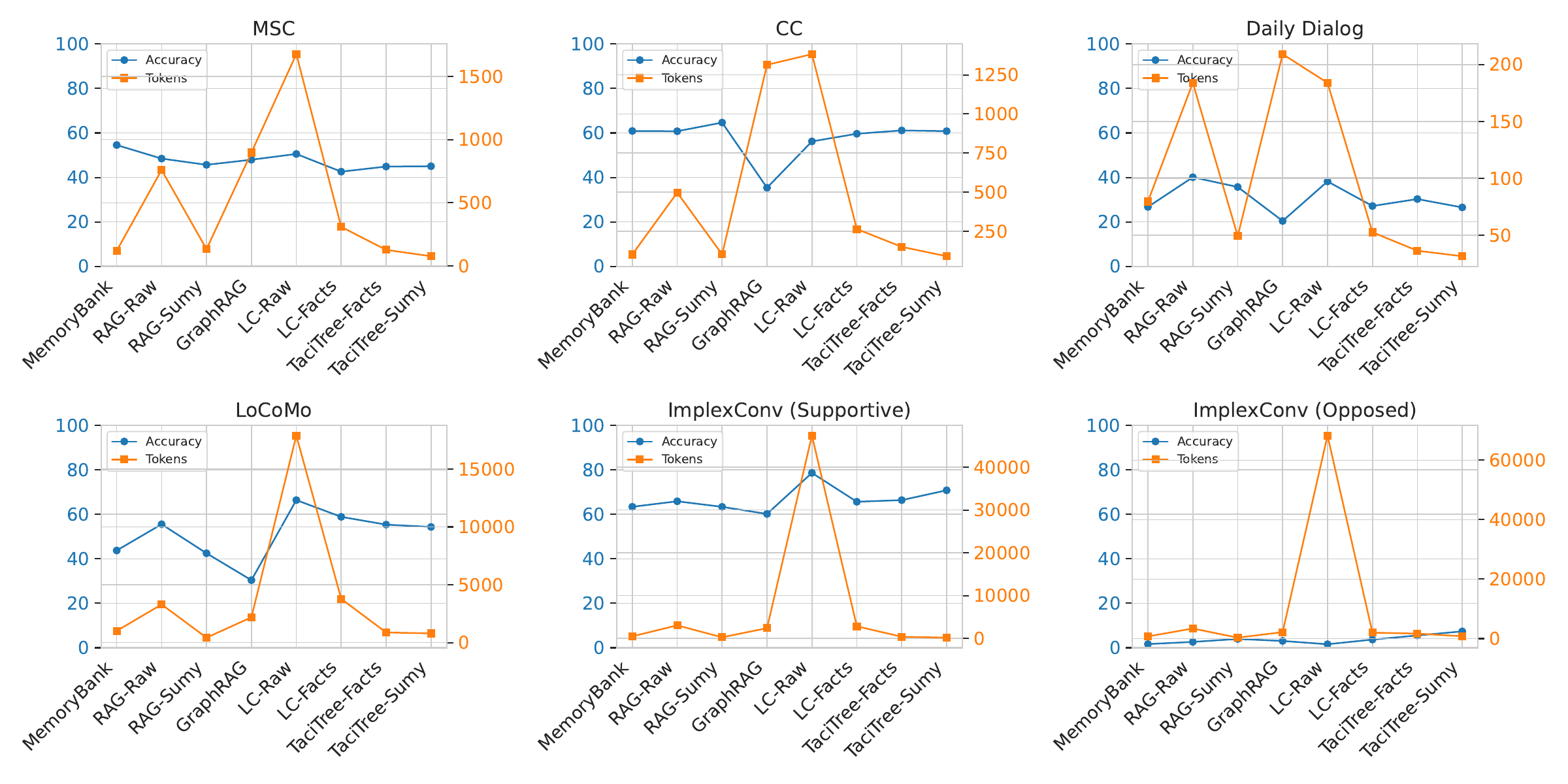}
    \vspace{-3mm}
  \caption{Response accuracy (blue) and retrieved token size (orange) across different frameworks and datasets.}
    \label{fig:acc_token}
    \vspace{-3mm}
\end{figure*}

\subsection{Information Retrieval Accuracy}
Our evaluation of retrieval accuracy, shown in Table~\ref{tab:retrieval_acc}, demonstrates that \ourmodel outperforms all baselines on \ourdata, achieving the highest F1 scores in both supportive (55.18\%) and opposed (14.84\%) implicit reasoning scenarios.
This highlights its ability to effectively retrieve relevant implicit knowledge, even when semantic similarity with the query is low. 
Traditional RAG-based approaches perform competitively on datasets without implicit reasoning (e.g., CC: 38.39\%, DailyDialog: 34.03\%), but their performance drops significantly on implicit reasoning tasks, with scores of 12.66\% (supportive) and 5.58\% (opposed), respectively. This indicates that standard retrieval techniques struggle with retrieving non-explicit evidence. 
While GraphRAG incorporates structured retrieval through graph-based knowledge representation, it still underperforms across all datasets, suggesting that graph-based retrieval alone is insufficient for handling implicit reasoning (further analysis provided in the Appendix~\ref{sec:graphrag}).

\subsection{Response Accuracy and Token Size}\label{sec:tokensize}
Response accuracy is closely tied to the amount of retrieved content—larger retrieval sizes generally improve accuracy. 
To further examine the trade-off between response accuracy and token efficiency, we analyze the performance of different frameworks across datasets.
As shown in Figure~\ref{fig:acc_token}, Table~\ref{tab:accuracy_results} and~\ref{tab:token_results}, models with larger token sizes tend to achieve higher QA accuracy, such as the long-context methods using raw conversations.
However, this comes at a significant computational cost. For instance, in LoCoMo, retrieving the entire conversation results in 17,896 tokens, yet the accuracy gain remains marginal compared to more efficient approaches.
Our framework, \ourmodel, strikes a balance between accuracy and retrieval efficiency. For instance, in Supportive \ourdata, our model achieves an accuracy of 70.8\% while retrieving only 172.66 tokens, compared to RAG-Raw, which retrieves 3045.94 tokens but only reaches 65.89\% accuracy. 
This shows that \ourmodel efficiently selects relevant information while maintaining strong performance.
Furthermore, in opposed reasoning cases, long-context methods perform poorly due to their tendency to retrieve excessive semantically relevant but misleading information. 
This reinforces the importance of selective retrieval in implicit reasoning tasks. 
Additionally, in datasets like MSC, CC, and DailyDialog, where conversations are human-human, responses often rely on personal context and may not require background retrieval.
Generated answers in these datasets may still be valid even when differing from the ground truth, we conduct a human evaluation (detailed in the Appendix~\ref{sec:humaneval}) for a fair assessment.

\begin{table}[t]
    \centering
    \small
    \begin{tabular}{lcccc}
        \toprule
        \multirow{2}{*}{\textbf{Model}} & \multicolumn{2}{c}{\textbf{\ourmodel}}  & \multicolumn{2}{c}{\textbf{Long-Context}}\\
        \cmidrule(lr){2-3} \cmidrule(lr){4-5} 
        & \textbf{Summary} & \textbf{Facts}  & \textbf{Raw} & \textbf{Facts} \\
        \midrule
        GPT-4o-mini  & 5.53 & 7.37 & 2.42 & 6.84 \\
        GPT-o3-mini & 9.97 & 9.90  & 2.42 & 6.84 \\
        GPT-o1 & 29.70 & 28.71  & 4.95 & 21.78 \\
        \bottomrule
    \end{tabular}
    \caption{Response accuracy across different models.}
    \vspace{-3mm}
    \label{tab:response_acc}
    
\end{table}

\subsection{Opposed Implicit Reasoning Analysis}
Table~\ref{tab:response_acc} presents the results for opposed reasoning, where retrieval accuracy remains high, but response accuracy is notably low. 
This suggests that implicit reasoning is particularly challenging, as noisy yet lexically relevant conversations can obscure the correct answer. 
To address this, we tested more powerful LLMs, including GPT-o3-mini and GPT-o1. The results indicate a clear trend: more powerful models perform better, with GPT-o1 achieving the highest accuracy at 29.70\%, while GPT-o3-mini also outperforms GPT-4o-mini. 
These findings highlight the importance of stronger reasoning capabilities in handling complex implicit reasoning tasks.

\section{Conclusion and Future Work}
In this work, we introduce \ourdata, a large-scale multi-session dataset designed to evaluate implicit reasoning in long-term personalized conversations. 
Unlike existing benchmarks, \ourdata incorporates subtle, semantically distant reasoning patterns that challenge traditional retrieval and long-context modeling approaches. 
To address these challenges, we propose \ourmodel, a hierarchical tree-based framework that efficiently retrieves implicit knowledge while maintaining token efficiency. Our experiments demonstrate that IMPACT significantly improves retrieval and response accuracy, outperforming baselines with less cost. 
Future work includes enhancing implicit reasoning capabilities by integrating adaptive retrieval mechanisms and exploring more advanced LLM architectures to better handle complex, context-dependent reasoning in long-term dialogues.

\section*{Limitations}
Some conversations in \ourdata are synthetically generated through LLM prompting, making them inherently sensitive to the specific prompts used. Since both data generation and content retrieval rely on LLM responses, the results can vary based on prompt phrasing and model updates, leading to potential inconsistencies across runs. Although we carefully design prompts to ensure high-quality reasoning scenarios, maintaining strict reproducibility remains a challenge.

\bibliography{main}

\appendix

\section{Related Work}\label{sec:relatedwork}
\paragraph{Long-term or Multi-session Dialogue Datasets}
Existing long-term dialogue datasets focus on different aspects of conversational memory and personalization. 
MSC\cite{xu2021beyond} introduces five-session human-human dialogues with annotated summaries to enhance continuity. 
CC\cite{jang2023conversation} and LoCoMo\cite{maharana2024evaluating} generate long-term multi-modal conversations with structured persona timelines. 
PerLTQA\cite{du2024perltqa} emphasizes personalized long-term QA without multi-session dialogue dynamics. 
Furthermore, DialSim\cite{kim2024dialsim} assesses dialogue similarity, focusing on coherence rather than long-term retrieval. Recent dataset \textsc{LongMemEval}\cite{wu2024longmemeval} tests memory retention over multi-turn dialogues but lacks personalization. In contrast, \ourdata (Table~\ref{tab:dataset_comparison}) is the first large-scale dataset designed for implicit reasoning in long-term conversations, incorporating both opposed and supportive scenarios that challenge retrieval-based and long-context models.

\paragraph{Long-term Memory Methodology}
To enhance long-term conversational reasoning, methods like MemoryBank~\cite{zhong2024memorybank} and LDAgent~\cite{li2024hello} incorporate structured memory mechanisms, while SCM~\cite{wang2023enhancing} utilizes structured conversational memory for efficient information retention and retrieval. 
RAG frameworks, including LlamaIndex, LangChain, and Haystack, enable structured retrieval to integrate relevant past context, with GraphRAG~\cite{edge2024local} further leveraging graph-based knowledge representation for improved contextual retrieval. 
Long-context processing remains an active research area, focusing on adapting LLMs to handle extended prompts; however, performance typically degrades as context length increases. Techniques such as hierarchical memory representations and adaptive retrieval mechanisms attempt to address this limitation. 
Our proposed approach, \ourmodel, introduces a hierarchical tree framework that structures conversation history into multiple levels of summarization, significantly improving LLMs’ ability to reason over long-term conversations with implicit contextual dependencies.

\section{Persona Extraction}\label{sec:persona}
We include some personas used to generate implicit conversations, such as, ``This person enjoys listening to pop music,'' ``This person likely engages in nostalgic experiences related to Azerbaijani culture,'' and ``This person is a casual listener of classic rock music.''All personas are formatted as short sample sentences, making it easier to create implicit reasoning.

\section{GraphRAG Analysis}\label{sec:graphrag}

Graph-based Retrieval-Augmented Generation (GraphRAG) extends traditional retrieval methods by structuring knowledge in a graph representation where nodes correspond to relevant entities or concepts, and edges encode relationships. This enables contextual retrieval beyond simple lexical similarity. However, in the case of our dataset, GraphRAG underperforms due to challenges in implicit reasoning, graph construction, and query generation.

The performance of GraphRAG heavily depends on the quality of the graph it constructs. In traditional explicit reasoning tasks, nodes represent well-defined entities (e.g., named entities, known facts) and edges reflect structured relationships. However, our dataset focuses on implicit reasoning, where relevant connections are syntactic, semantic, or pragmatically inferred rather than explicitly defined. As a result, the model struggles to create meaningful edges that capture indirect relationships between persona traits. Key implicit details are often embedded across multiple dialogue sessions, making single-instance graph representations insufficient. Graph sparsity leads to retrieval failures, as distant yet relevant information remains inaccessible due to missing edges.

Effective retrieval in GraphRAG depends on correctly structuring queries that retrieve the most relevant nodes. However, in our dataset, implicit reasoning requires multi-hop retrieval, yet GraphRAG often retrieves single-hop neighbors, missing deeper contextual connections. Furthermore, the LLM relies on semantic similarity-based retrieval, which fails when implicit reasoning requires retrieving conceptually related but lexically distant nodes. In addition, over-reliance on direct lexical matching leads to retrieval noise, where GraphRAG incorrectly prioritizes surface-level matches over deeper reasoning-based connections.

% Prompt tuning is crucial to mitigate these issues. We explored various approaches to improve retrieval accuracy:
% \paragraph{Enhanced Query Structuring:} Instead of allowing the LLM to directly retrieve nodes, we introduce an intermediate step where the model first identifies key reasoning elements before constructing a query. This structured approach improves the quality of retrieved information.

% \paragraph{Edge Selection Refinement:} We modify the retrieval prompt to encourage the model to justify its edge selection. This helps filter out irrelevant nodes and refocus retrieval on meaningful, implicit connections.

% \paragraph{Rate-Limited Document Processing:} Our implementation incorporated a rate-limited document filtering step, ensuring that each retrieved document undergoes LLM-driven graph transformation before addition to the knowledge graph. This helped in refining the graph's structure to be more aligned with implicit reasoning queries.

% \paragraph{Schema-Guided Cypher Query Generation:} We employed a structured Cypher generation prompt that constrains retrievals based on explicit schema information, ensuring that generated queries follow a hierarchical reasoning structure.

\begin{table}[t]
    \centering
    \small
    \begin{tabular}{lcccc}
        \toprule
        \multirow{2}{*}{\textbf{Model}} & \multicolumn{2}{c}{\textbf{MSC}}  & \multicolumn{2}{c}{\textbf{DailyDialog}}\\
        \cmidrule(lr){2-3} \cmidrule(lr){4-5} 
        & \textbf{LLM} & \textbf{Human}  & \textbf{LLM} & \textbf{Human} \\
        \midrule
        \ourmodel  & 45.03 & \textbf{55.18} & 26.51 & \textbf{47.06} \\
        MemoryBank & 54.57 & \textbf{64.30}  & 26.73 & \textbf{53.27} \\
        RAG & 45.67 & \textbf{53.24}  & 40.08 & \textbf{60.15} \\
        \bottomrule
    \end{tabular}
    \caption{Comparison between Human and LLM Evaluations.}
    \vspace{-3mm}
    \label{tab:human_evl}
    
\end{table}

\section{Human Evaluation}\label{sec:humaneval}
Table~\ref{tab:human_evl} presents a comparison of human and LLM-based evaluations across different models on the MSC and DailyDialog datasets. 
In all cases, human evaluation scores are higher than LLM evaluation scores. This suggests that even when a model's generated response does not match the ground truth, it can still be considered a valid answer to the given question. 
This is reasonable since MSC and DailyDialog consist of human-to-human conversations, where responses can often be personal. 
For example, beyond simply answering a question, a response may include follow-up questions or introduce new topics, reflecting natural conversational dynamics.

\begin{table*}[h]
    \small
    \centering
    \renewcommand{\arraystretch}{1.2}
    \resizebox{\linewidth}{!}{
    \begin{tabular}{llcccccc}
        \toprule
        \multirow{2}{*}{\textbf{Category}} & \multirow{2}{*}{\textbf{Method}} & \multicolumn{6}{c}{\textbf{Datasets}} \\
        \cmidrule(lr){3-8}
        & & \textbf{MSC} & \textbf{CC} & \textbf{Daily Dialog} & \textbf{LoCoMo} & \textbf{IMPersona (Supp.)} & \textbf{IMPersona (Opp.)} \\
        \midrule
        Memory-based
        & MemoryBank & \textbf{54.57} & 60.89 & 26.73  & 43.75 & 63.41 & 1.65 \\
        \midrule
        \multirow{3}{*}{RAG} 
        & Raw  & 48.47 & 60.78 & \textbf{40.08} & 55.56 & 65.89 & 2.63 \\
        & Summary & 45.67 & \textbf{64.67} & 35.70 & 42.51 & 63.38 & 3.95 \\
        & GraphRAG & 47.96 & 35.35 & 20.41 & 30.40 & 60.20 & 3.06 \\
        \midrule
        \multirow{2}{*}{Long-Context} 
        & Raw & 50.55 & 56.20 & 38.20 & \textbf{66.42} & \textbf{78.66} & 1.58 \\
        & Facts & 42.59 & 59.64 & 27.14 & 58.89 & 65.68 & 3.68 \\
        \midrule
        \multirow{2}{*}{IMPACT} 
        & Facts & 44.89 & 61.13 & 30.27 & 55.39 & 66.41 & 5.53 \\
        & Summary & 45.03 & 60.81 & 26.51 & 54.35 & 70.83 & \textbf{7.37} \\
        \bottomrule
    \end{tabular}
    }
    \caption{Response Accuracy across Different Frameworks and Datasets}
    \label{tab:accuracy_results}
\end{table*}
\begin{table*}[h]
    \small
    \centering
    \renewcommand{\arraystretch}{1.2}
    \resizebox{\linewidth}{!}{
    \begin{tabular}{llcccccc}
       \toprule
        \multirow{2}{*}{\textbf{Category}} & \multirow{2}{*}{\textbf{Method}} & \multicolumn{6}{c}{\textbf{Datasets}} \\
        \cmidrule(lr){3-8}
        & & \textbf{MSC} & \textbf{CC} & \textbf{Daily Dialog} & \textbf{LoCoMo} & \textbf{IMPersona (Supp.)} & \textbf{IMPersona (Opp.)} \\
        \midrule
        Memory-based
        & MemoryBank & 120.22 & 100.55 & 79.49 & 1015.10 & 479.34 & 743.44 \\
        \midrule
        \multirow{3}{*}{RAG} 
        & Raw      & 759.26 & 496.39 & 183.86 & 3296.15 & 3045.94 & 3395.81 \\
        & Summary & 133.41 & 101.17 & 49.10 & \textbf{426.17} & 262.43 & \textbf{319.33} \\
        & GraphRAG & 896.78 & 1314.37 & 209.08 & 2182.91 & 2394.82 & 2178.09 \\
        \midrule
        \multirow{2}{*}{Long-Context} 
        & Raw & 1675.04 & 1382.94 & 183.86 & 17896.13 & 47384.54 & 68299.85 \\
        & Facts & 309.23 & 263.11 & 52.49 & 3778.62 & 2825.89 & 2002.24 \\
        \midrule
        \multirow{2}{*}{IMPACT} 
        & Facts & 127.61 & 149.22 & 36.28 & 877.98 & 384.21 & 1682.26 \\
        & Summary & \textbf{77.07} & \textbf{89.06}  & \textbf{31.42} & 791.48 & \textbf{172.66} & 786.97 \\
        \bottomrule
    \end{tabular}
    }
    \caption{Average Number of Tokens across Different Frameworks and Datasets}
    \label{tab:token_results}
\end{table*}

\begin{figure*}[t]
\begin{minipage}{1.0\textwidth}
\begin{lstlisting}[language=Python, label=dom_ppt, caption={Prompt to extract persona traits}]
prompt = "
   Here is a brief description of a person:
    {persona}
    
    Please break it down into several components, including: "demographics" (including name, age, living location, birthplace, marital status, etc.), "career_life_and_goals" (make sure this part only contains things related with the person's career life), and "everyday_life_and_hobbies" (make sure this part is nothing related with the person's career life). Just list those information that are presented and leave others that are unknown. Below are some examples, try to make each point separate from each other and self-explanable. Only output a JSON object like in the following examples.
    Example 1:
    Input: An eco-friendly lifestyle podcaster who features change-makers and promotes sustainable living
    Output:
    ```json
    {{
        "demographics": {{
            "occupation": "This person is an eco-friendly lifestyle podcaster."
        }},
        "career_life_and_goals": [
            "This person features change-makers and promotes sustainable living."
        ]
    }}
    ```
    
    Example 2:
    Input: a nostalgic Azerbaijani pop music lover
    Output:
    ```json
    {{
        "demographics": {{
            "nationality": "This person is from Azerbaijani."
        }},
        "everyday_life_and_hobbies": [
            "This per``son enjoys listening to pop music.",
            "This person likely engages in nostalgic experiences related to Azerbaijani culture."
        ]
    }}
    ```
    "

\end{lstlisting}
\end{minipage}
\end{figure*}

\begin{figure*}[t]
\begin{minipage}{1.0\textwidth}
\begin{lstlisting}[language=Python, label=dom_ppt, caption={Prompt to generate opposed implicit reasoning}]
prompt = "
    {per_info} However, they have not been able to do it recently. Can you give me at least 20 implicit reasons why that person cannot do it? 
    The reasons should be completely different from each other and belong to different categories.
    The reason should be specific with detailed information, like why it happens.
    The reason cannot include words related to "{traits_info}"
    Please explain the reasoning in only one sentence. Please only output the reasons with the format:
    1: 
    2: 
    "

\end{lstlisting}
\end{minipage}
\end{figure*}

\begin{figure*}[t]
\begin{minipage}{1.0\textwidth}
\begin{lstlisting}[language=Python, label=dom_ppt, caption={Prompt to generate supportive implicit reasoning}]
prompt = "
    {per_info} Can you give me at least 20 implicit reason information that supports this claim? Therefore, if I ask you, "Does {per_info}?", you have to answer "yes". 
    The reason information should be completely different from each other and belong to different categories.
    The reason should be specific with detailed information, like why it happens.
    The reason cannot include words related to "{traits_info}"
    Please explain the reasoning in only one sentence. Please only output the reasons with the format:
    1: 
    2: 
    "

\end{lstlisting}
\end{minipage}
\end{figure*}

\begin{figure*}[t]
\begin{minipage}{1.0\textwidth}
\begin{lstlisting}[language=Python, label=dom_ppt, caption={Prompt to generate opposed implicit question}]
prompt = "
    Here's the conversation between a user(speaker 1) and a chatbot assistant.
    Speaker 1 has the following persona trait: {per_info}. However, speaker 1 cannot do the trait due to the reason that {reason_info}. 
    Now, speaker 1 asks you a question related to the trait. {reason_info} affect your answer to this question.
    You should tell speaker 1 they cannot do the trait due to the reason.
    The trait should be mentioned in the question.
    The question itself should not mention the reason or effect of the reason.
    Questions should be asked in the first person. Include "I".
    The question should not be a yes/no question. 
    The question needs to be diverse.

    Please only output the question in the format of less than 20 words without any additional sentences.
    "

\end{lstlisting}
\end{minipage}
\end{figure*}

\begin{figure*}[t]
\begin{minipage}{1.0\textwidth}
\begin{lstlisting}[language=Python, label=dom_ppt, caption={Prompt to select opposed implicit reasoning}]
prompt = "
    {per_info}. Here are potential implicit reasons why this person is unable to follow this trait: {str_reason}. 
    Could you select the reason that is both the most logically sound and subtly implied?
    Please select only from the provided options and output the reason only.
    "

\end{lstlisting}
\end{minipage}
\end{figure*}

\begin{figure*}[t]
\begin{minipage}{1.0\textwidth}
\begin{lstlisting}[language=Python, label=dom_ppt, caption={Prompt to generate noisy scenarios}]
prompt = "
    Consider a person with specific personality traits {persona} that could serve as responses to a given question {question}. 
    Can you generate additional scenarios that reflect or align with these personality traits to support the question?
    Please output 5 scenarios that are relevant to the given traits and question.
    The scenarios should contain only one sentence.
    The scenarios can talk about both {traits_info} or other stuff that is related to {traits_info} but do not have to be the same.
    Please output the scenarios only with the index number.

    For example:

    Trait: I love sports
    Question: I'm bored; can you give me some suggestions?
    Scenarios:
    1. I love playing basketball.
    2. My favorite basketball player is Stephen Curry.
    "

\end{lstlisting}
\end{minipage}
\end{figure*}

\begin{figure*}[t]
\begin{minipage}{1.0\textwidth}
\begin{lstlisting}[language=Python, label=dom_ppt, caption={Prompt to generate noisy conversations}]
prompt = "
    There are two speakers. Speaker 1 encounters the scenario that "{scenario}". Speaker 2 is the AI assistant.
    Based on the information. Can you generate a conversation with at least 10 turns?
    Speaker 1 shouldn't mention the scenario too early. It must be mentioned in the later section.
    Speaker 1 is exactly the person who encounters the scenario.
    The beginning turns should serve as a warm-up to introduce the scenario in a natural way.
    The conversation should be centered around the scenario without any irrelevant or extra information that is not related to the scenario.
    For Spearker 1, please do not start the conversation by saying something similar to "I'm feeling a bit overwhelmed lately." or use the same format as this sentence.
    Include diverse styles like detailed explanations, step-by-step guidance, casual small talk, humor, storytelling, and problem-solving. 
    The conversation should feel realistic and flow naturally. 
    Aim for a balance of formality and informality, capturing nuanced exchanges that go beyond simple responses.
    Please output the conversation in the following format:
    Speaker1: ...
    Assistant: ...
    
    Speaker1: ...
    Assistant: ...
    "

\end{lstlisting}
\end{minipage}
\end{figure*}

\begin{figure*}[t]
\begin{minipage}{1.0\textwidth}
\begin{lstlisting}[language=Python, label=dom_ppt, caption={Prompt to summarize facts}]
prompt = "
    Can you summarize {text} in one sentence to only contain the high-level information? 
    Please only output the summary without anything else.
    "

\end{lstlisting}
\end{minipage}
\end{figure*}

\end{document}